\documentclass{article}

\PassOptionsToPackage{numbers, compress}{natbib}

\usepackage[preprint]{neurips_2019}

\usepackage[utf8]{inputenc} 
\usepackage[T1]{fontenc}    
\usepackage{hyperref}       
\usepackage{url}            
\usepackage{booktabs}       
\usepackage{amsfonts}       
\usepackage{nicefrac}       
\usepackage{microtype}      
\usepackage{graphicx}
\usepackage{amsmath}
\usepackage{bm}
\usepackage{floatrow}
\newfloatcommand{capbtabbox}{table}[][\FBwidth]
\title{Hybrid Low-order and Higher-order Graph Convolutional Networks}

\author{%
  FangYuan~Lei$^{1}$ \And Xun~Liu$^{1}$ \And QingYun~Dai$^{1}$ \And Bingo Wing-Kuen~Ling$^{2}$\\
  \AND
  Huimin~Zhao$^{3}$ \And Yan~Liu$^{1}$ \and\\
  $^{1}$School of Electronic and Information, Guangdong Polytechnic Normal University, Guangdong, China\\
  $^{2}$School of Information Engineering, Guangdong University of Technology, Guangdong, China\\
  $^{3}$School of Computer Sciences, Guangdong Polytechnic Normal University, Guangdong, China\\
  \texttt{\{leify\}@126.com,  \{1834720998, 1144295091\}@qq.com} \\
  \texttt{\{yongquanling\}@gdut.edu.cn,  \{zhaohuimin\}@gpnu.edu.cn,  \{27890726\}@qq.com} \\
}

\begin{document}

\maketitle

\begin{abstract}
  With higher-order neighborhood information of graph network, the accuracy of graph representation learning classification can be significantly improved. However, the current higher order graph convolutional network has a large number of parameters and high computational complexity. Therefore, we propose a Hybrid Lower order and Higher order Graph convolutional networks (HLHG) learning model, which uses weight sharing mechanism to reduce the number of network parameters. To reduce computational complexity, we propose a novel fusion pooling layer to combine the neighborhood information of high order and low order. Theoretically, we compare the model complexity of the proposed model with the other state-of-the-art model. Experimentally, we verify the proposed model on the large-scale text network datasets by supervised learning, and on the citation network datasets by semi-supervised learning. The experimental results show that the proposed model achieves highest classification accuracy with a small set of trainable weight parameters.
\end{abstract}

\section{INTRODUCTION}

Convolutional neural networks (CNNs) have achieved great success in grid structure data such as image and video~\cite{krizhevsky2012imagenet,he2016deep}. It is attributed to a series of filters of convolutional layers from the CNNs which can obtain local invariant features. Compared to a regularized network, the number of neighbors of a node in the graph network may be different. Therefore, it is difficult to implement the filter operator directly on an irregular network structure~\cite{defferrard2016convolutional,li2018deeper}.

In the graph network, the nodes and the connecting edges between them contain abundant network characteristic information. Graph convolutional network (GCN) adopts aggregation of neighborhood nodes to realize continuous information transmission based on graph network~\cite{gilmer2017neural}. By making full use of this information, GCN can effectively achieve tasks such as classification, prediction and recommendation~\cite{kipf2016semi,hamilton2017inductive,yao2018graph}.

Bruna et al.~\cite{bruna2013spectral} and Cao et al.~\cite{cao2015grarep} applied a generalized convolutional network to the graph frequency domain by the Fourier Transformation. In this method, eigenvalue decomposition is performed on the neighborhood matrix. To reduce  computational complexity, Defferrard et al.~\cite{defferrard2016convolutional} proposed the Chebyshev polynomials to achieve local graph convolution. Kipf and Welling~\cite{kipf2016semi} proposed a classical GCN, which was approximated by a first-order Chebyshev polynomial. This approach reduces computational complexity but introduces truncation errors. This will result in the inability to capture high-level interaction information between nodes in the graph, and also limits the capabilities of the model. The propagation process of information in the graph is not only related to its first order neighborhood, but also to its higher order neighborhood. Therefore, the rational use of second order neighborhoods, third order neighborhoods and other high order neighborhood information will be beneficial to target classification prediction accuracy~\cite{Lee2018Higher,abu2018a,abu2019mixhop,Ma2018SimilarityLW}.

Based on the classical GCN~\cite{kipf2016semi}, to make full use of high order and low order neighborhood information, we propose a novel Hybrid Low order and Higher order Graph convolutional network (HLHG). As shown in Figure 1, the graph convolutional layer of our model is a simple and effective to capture high order neighborhoods information, and nonlinear fuse different order neighborhood information. The contributions are summarized as follows:

1) We propose a new fusion pooling layer to achieve high order neighborhood fusion with the low order neighborhood of graph networks.

2) We propose a low order neighborhood and high order neighborhood weights sharing mechanism to reduce the computational complexity and parameter quantities of the model.

3) The experimental results show that our HLHG achieves the state-of-the-art in both the text network classification of supervised learning and the citation network of semi-supervised learning.

\begin{figure}[h]
\begin{center}
\includegraphics[width=\textwidth]{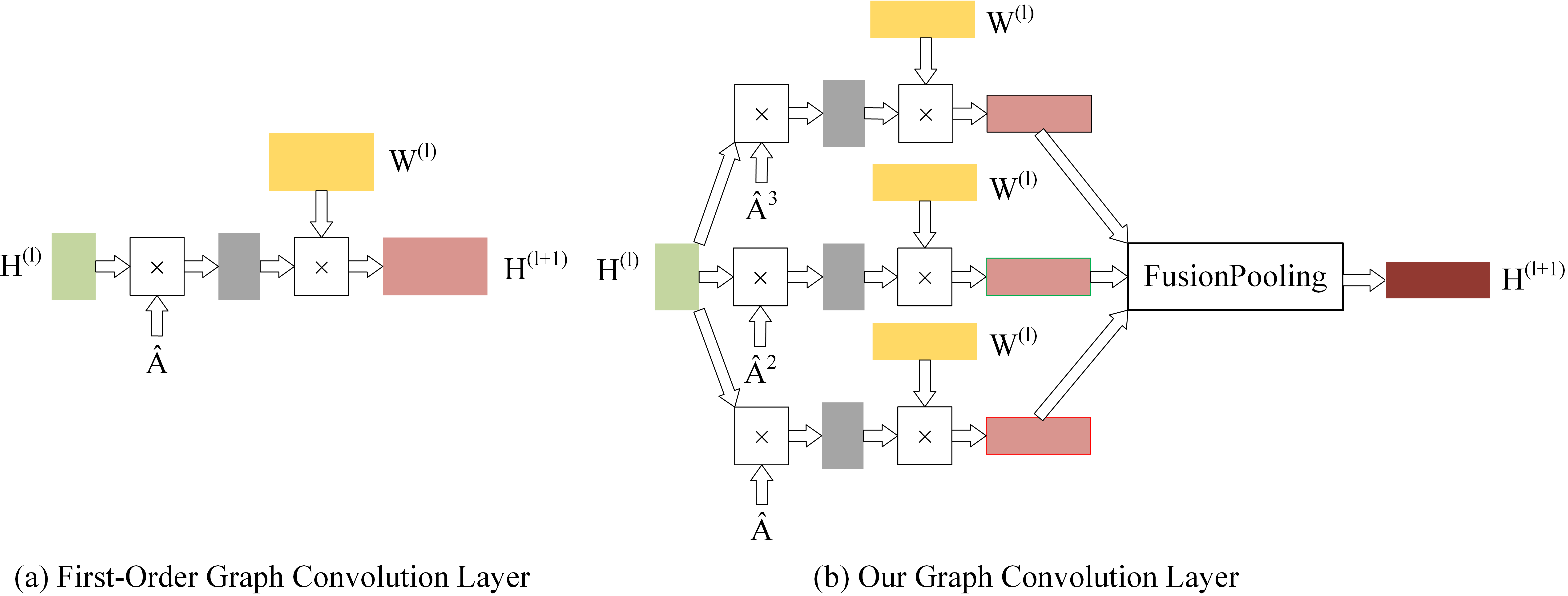}
\caption{The Graph Convolutional Layer of Our Model. (a) is the First Order Graph Convolutional Layer of the Kipf et al.~\cite{kipf2016semi} model. The input value is $ H^{(l)} $, and output is $ H^{(l+1)} $, and the trainable parameter is $ W^{(l)} $. (b) The Third Order Graph Convolutional Layer of our HLHG model. Different order neighborhood matrix shares the trainable weight.}
\label{figure/fig1}
\end{center}
\end{figure}

\section{Related Work}
\subsection{Graph Convolutional Network}

Given a graph $ G=(V,E) $, where $ V $ is the set of nodes and $ E $ is the set of edges. If the node $ V_{i} $ and $ V_{j} $ connect, then $ E_{ij}=1 $, otherwise $ E_{ij}=0 $. The information in the graph propagates along with the edges $ E $. It also applies when considering the network nodes self-loop, which means $ E_{ii}=1 $. Assuming that the information propagates by each node in the graph network is $ x\in R^{r} $, the information matrix in the graph is $ X\in R^{n\times r} $, where the $ n $ is the total number of the node in the graph network. And the $ r $ is the dimension of information feature. It assumes that the loop graph network $ G $ represents as $ \tilde{G} $, then the adjacency matrix of the graph network $ \tilde{G} $ is represented as $ \tilde{A}=(A+I) $. The degree matrix of $ \tilde{A} $ in the graph network $ \tilde{G} $ is the diagonal matrix, $ \tilde{D}_{ii}=\sum_{j}\tilde{A}_{ij} $.

In the spectral domain, Bruna et al.~\cite{bruna2013spectral} generalized the convolutional network to the graph network by graph Fourier transform. And the convolutional filter is extended to the frequency pattern through aggregating the neighborhood nodes of the object node. This transformation process involves the eigenvalue decomposition of the Laplacian matrix of the graph network, which costs expensive resource for large scale graph network.

Kipf and Welling~\cite{kipf2016semi} proposed the classical graph convolutional neural network model based on the Fourier transform. To achieve efficient and localized filters, the spectral filters are parameterized as Chebyshev polynomials of eigenvalues of the graph Laplacian to reduce the computation burden. Therefore, the spectral filters rely on the spectrum of the graph Laplacian. The GCN model approximates the model using a first-order Chebyshev polynomial. The propagation model in the graph network is:
\begin{equation}
H^{(l+1)}=\sigma(\hat{A}H^{(l)}W^{(l)}) \,,
\label{eq1}
\end{equation}
where $ \hat{A}=\tilde{D}^{-\frac{1}{2}}\tilde{A}\tilde{D}^{-\frac{1}{2}}\in R^{n\times n} $ is the regularized adjacency matrix. $ H^{(l)} $ denotes the information propagate matrix, and $ W^{(l)} $ represents as the trainable weight of layer $ l $. When $ l=1 $, $ H^{(1)}=X\in R^{n\times r} $, which means the initial input value of the GCN. $ \sigma(\cdot) $ denotes the activation function. In order to reduce the computational complexity, the convolution operator in the graph is defined by a simple neighborhood average. However, the convolutional filters are too simple to capture high-level interaction information between nodes in the graph. Therefore, the classification accuracy on citation network datasets is little low.

\subsection{High order Graph Convolutional Network}

In the graph network, the node information propagation along edge, which not only relates to its first-order neighborhood, but also to its higher-order neighborhood.

Lee et al.~\cite{Lee2018Higher} propose the Motif Convolutional Network (MCN), which aggregates the first-order neighborhood information of the vertices in the graph and its high-order Motif information. Ma et al.~\cite{Ma2018SimilarityLW} propose the high-order convolution which characterizes and learns the community structure in the graph network by combining high-order neighborhoods in the graph convolutional network.

Abu-El-Haija et al.~\cite{abu2018a,abu2019mixhop} propose a high order graph convolution model based on GCN model. The propagation model of the high-order graph convolution is as shown in the Eq.~(\ref{eq2}). In this model, the transfer function of the $(l+1)$th layer is a column concatenation from the first order to $ p $th order in the $l$-th layer. In the propagation model, different order neighborhoods of the same layer use different weight parameters.
\begin{equation}
H^{(l+1)}=\sigma(B^{1}H^{(l)}W_{1}^{(l)}|B^{2}H^{(l)}W_{2}^{(l)}|\cdot\cdot\cdot|B^{p}H^{(l)}W_{p}^{(l)}) \,,
\label{eq2}
\end{equation}
where $ B=\tilde{D}^{-\frac{1}{2}}\tilde{A}\tilde{D}^{-\frac{1}{2}} $. However, with the network layers deeper, the dimensions of the $ H^{(l+1)} $ will increase which propagate between layers. Therefore, the trainable weight parameters will be more, and the training resource will also increase.

\section{Method}

When the message pass through the graph network, the nodes will receive latent representations along from their first-hop nodes and also from the N-hops neighbors every time. In this section, we propose a model to aggregate the trainable parameters which can choose how to mix latent message from various hops nodes.

\subsection{General Information Fusion Pooling}

The information propagation of the graph network is passed along the edges between the vertices in the graph. It assumes that the graph network $ G=(V,E) $ is an undirected graph. The general procedure of fusion pooling is described as follows. It assumes that the $ k $-th neighborhood $ A^{(k)}=[a_{ij}^{k}] $, the result after fusion pooling operator $ Z^{(k)}=[Z_{ij}^{k}] $, where $ z_{ij}^{k}=max(a_{ij}^{1},a_{ij}^{2},\cdot\cdot\cdot,a_{ij}^{k}) $, where $ k $ is the hop from the given node.

There has an example to show how to fuse the different order neighborhood. For a given adjacency matrix $ \hat{B} $, it assumes that the $ h_{0} $ denote the first order neighborhood, and and $ h_{1} $ denote the second order neighborhood.

If $ h_{0}=\hat{B}XW_{1}=\begin{bmatrix} 1 & 0 \\ 1 & 1 \\ \end{bmatrix} $ and $ h_{1}=\hat{B}^{2}XW_{1}=\begin{bmatrix} -1 & 0 \\ 2 & 1 \\ \end{bmatrix} $, then $ Pmax=(h_{0},h_{1})=\begin{bmatrix} 1 & 0 \\ 2 & 1 \\ \end{bmatrix} $.

In the process of information dissemination and fusion, the first-order neighborhood features are fully considered, and high-order neighborhood features are also considered. Therefore, the classification accuracy should be improved.

\subsection{Our Proposed Model}

\begin{figure}[h]
\begin{center}
\includegraphics[width=\textwidth]{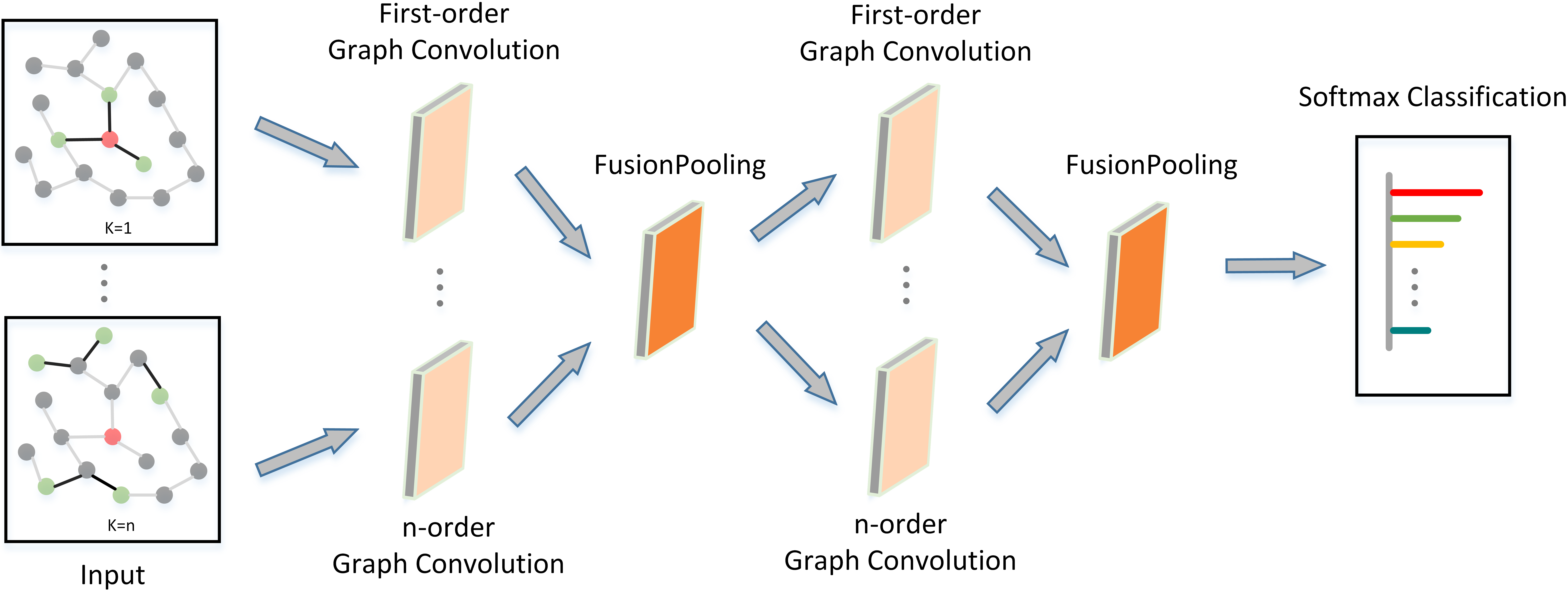}
\caption{Our HLHG model. The graph convolutional network layer of the HLHG model consists of two layers, and the information fusion pooling. The input parameters are first order to n order neighborhoods.}
\label{figure/fig2}
\end{center}
\end{figure}

In Figure~\ref{figure/fig2}, we propose the high order graph convolutional network model to fuse high order message which passes through the graph network. The model consists of an input layer, two graph convolutional layers, and an information fusion pooling layer connected to the graph convolutional layer. The softmax function is used for multi-classification output.

The model in this paper is to extend the classical GCN model~\cite{kipf2016semi} to the graph neural network model of higher-order neighborhoods. Each node in the model can get its representation from its neighborhood get and integrate messages. The system model is as follows:

\begin{equation}
Y=softmax[Pmax(\hat{A}\sigma(H^{(l)})W_{l},\cdot\cdot\cdot,\hat{A}^{(p)}H^{(l)})W_{l}] \,,
\label{eq3}
\end{equation}

where $ p $ is the order of the neighborhood, $ \hat{A}^{(p)}=\hat{A}^{(p-1)}\hat{A} $. $ \sigma(\cdot) $ is the activation function. $ W_{l} $ is the trainable weight parameter of $ l $ layer in the graph network. $ Pmax(\cdot) $ represents hybrid high order and low order of the information fusion. When parameter $ l=1 $, $ H^{(2)}=max(\hat{A}H^{(1)}W_{1},\cdot\cdot\cdot,\hat{A}^{(p)}H^{(1)}W_{1}) $, which is output of the first convolutional layer of the graph propagation model. And $ H^{(1)}=X\in R^{n\times r} $, which represents the initial input of our model.

In the preliminary experiment, we found that the two-layer high and low order mixed graph convolution is better than the one-level high and low order mixed graph convolution, and stacking more layers does not significantly improve the accuracy of the graph recognition task. Therefore, this paper uses a 2-layer graph convolution layer. In further experiments, we validate the $ p=2 $ and $ p=3 $ in Eq.~(\ref{eq3}) for our HLHG models. In the classification tasks of supervised learning and unsupervised learning, our HLHG models show very good performance between classification accuracy and computational complexity. We also validate $ p=4 $ or higher, the classification accuracy is not significantly improved. Therefore, we only analyze and implement our model in cases of $ p=2 $ and $ p=3 $ in the following sections.

In the Eq.~(\ref{eq3}), the model with $ p=2 $, that is, the hybrid model of the 1st and 2nd order neighborhoods is called the HLHG-2 model. The model $ p=3 $, that is, the hybrid model of the 1st, 2nd, and 3rd order neighborhoods is called the HLHG-3 model.

In the HLHG-2 model, it assumes that the graph convolutional network has 2 convolutional layers and the activation function is Relu. Then the output $ Y $ of the HLHG-2 model can be expressed as:

\begin{equation}
Y=softmax[Pmax[\hat{A}(Relu(M2))W_{2},\hat{A}^{2}(Relu(M2))W_{2}]] \,,
\label{eq4}
\end{equation}

where $ M2=Pmax(\hat{A}XW_{1},\hat{A}^{2}XW_{1}) $, and $ Pmax $ denotes the FusionPooling.

The same as HLHG-2 model, the output $ Y $ of the HLHG-3 model can be expressed as:

\begin{equation}
Y=softmax[Pmax[\hat{A}(Relu(M3))W_{2},\hat{A}^{2}(Relu(M3))W_{2},\hat{A}^{3}(Relu(M3))W_{2}]] \,,
\label{eq5}
\end{equation}

where $ M3=Pmax(\hat{A}XW_{1},\hat{A}^{2}XW_{1},\hat{A}^{3}XW_{1}) $.

For large scale graph network, it is unacceptable to directly calculate $ \hat{A}^{3}=\hat{A}^{2}\hat{A}=\hat{A}\hat{A}\hat{A} $. Therefore we take the $ \hat{A}^{3}XW_{1}=\hat{A}(\hat{A}(\hat{A}X))W_{1} $. This procedure will reduce the computation complexity.

Therefore, our HLHG model has 2-layer graph network, the iterative expression of the 2nd-order neighborhood is as follows:

\begin{equation}
Y=softmax(\hat{A}Relu(H)W_{2},\hat{A}^{2}Relu(H)W_{2}) \,,
\label{eq6}
\end{equation}

where $ H=Pmax(\hat{A}XW_{1},\hat{A}^{2}XW_{1}) $. We use the $ Pmax $ presents our fusion pooling operator which take the maximum value in the corresponding element.

We use multi-classified cross entropy as the loss function of our HLHG model, $ L=-\sum_{i}\tilde{y_{i}}\log(q_{i}) $, where $ \tilde{Y} $ is the labeled samples. The graph neural network trainable weights $ W_{1} $ and $ W_{2} $ are trained using gradient descent. In each training iteration, we perform batch gradient descent.

\subsection{Computational Complexity and Parameter Quantity}
In the large scale graph network, it is difficult to directly calculate $ \hat{A}^{(p)} $. To reduce the computational complexity, we calculate the $ \hat{A}^{(p)} $ by iterative solution~\cite{abu2018a}. For higher order, the right to left multiplication iterative procedure as $ \hat{A}^{(p)}H^{(l)}W_{l}=\hat{A}(\hat{A}^{(p-1)}H^{(l)})W_{l} $. For example, when $ p=2 $, $ \hat{A}^{(2)}H^{(l)}W_{l}=\hat{A}(\hat{A}X)W_{l} $. Due to the weight sharing in the same convolutional layer, with $ \hat{A}XW_{1}\in R^{n\times r_{1}} $, $ \hat{A}^{(2)}XW_{1}\in R^{n\times r_{1}} $, and $ \hat{A}^{(k)}XW_{1}\in R^{n\times r_{1}} $. Here, $ W_{1}\in R^{r\times r_{1}} $ ($ r_{1} $ filters) and $ W_{2}\in R^{r_{1}\times r_{2}} $ ($ r_{2} $ filters) are the weight matrix of our first and second layer. $ k $ denotes the order of a sparse matrix $ \hat{A} $ with $ m $ non-zero elements. If $ r_{l} $ is the number of hidden neurons of the $l$-th convolutional layer, then the computational complexity of the $l$-th convolutional layer is $ O (r_{l}\times p\times m \times r_{l-1}) $ in our HLHG model. And the quantity of trainable weight is $ O (r_{l}\times r_{l-1}) $. The computational complexity of our HLHG model is $ O(\sum_{l=1}^{j}(r_{l}\times p\times m \times r_{l-1})) $, and the total quantity of trainable parameter is $ O(\sum_{l=1}^{j}(r_{l}\times r_{l-1})) $. The parameter $ j $ denotes the number of convolutional layers. when $ i=1 $, $ r_{0} $ represents the feature dimensions of the datasets, $ r_{1} $ represents the number of neurons of the first convolutional layer. For all the datasets, $ r_{0} >> r_{1} $ , therefore we only consider the first convolutional layer when we compare the computational complexity and parameter quantity.

Compared to~\cite{kipf2016semi}, we set fewer filters to maintain similar computational complexity and the parameter the amount is less via weight sharing both low order and higher order convolutions.

\section{Experiment}
\label{gen_inst}

In order to verify that our HLHG model can be applied to supervised learning and semi-supervised learning. On the text network datasets, we compare our model with the state-of-the-art methods by supervised learning. On the citation network datasets, we compare our model with the state-of-the-art methods by semi-supervised learning. For all experiments, we construct a 2-layer graph convolutional network of our model using TensorFlow. The code and data will be available at github.

\subsection{supervised text network classification}

We conduct supervised learning on five benchmark text graph datasets to compare the classification accuracy of HLHG with graph convolutional neural and other deep learning approaches.

\paragraph{Datasets}The datasets are 20-Newsgroups (20NG), Ohsumed, R52 and R8 of Reuters 21578 and Movie Review (MR). These data sets are publicly available on the web and are widely used as test-verified data sets.

These benchmark text datasets processed by Yao et al.~\cite{yao2018graph}, and convert text datasets into graph network structure. And use pre-processing to construct the adjacency matrix of the graph network input and input parameters. And the data set is divided into a training data set and a test data set in the same way.

\begin{table*}
\begin{floatrow}
\capbtabbox{
 \begin{tabular}{lcccccc}
	\toprule
    Datasets & C & D & Tr & Te & N    \\
    \midrule
    R52     & 52 & 9100 & 6532 & 2568 & 17992     \\
    OH     & 23 & 7400 & 3357 & 4043 & 21557      \\
    20NG     & 20 & 18846 & 11314 & 7532 & 61603  \\
    R8     & 8 & 7674 & 5485 & 2189 & 15362  \\
    MR     & 2 & 10662 & 7108 & 3554 & 29426  \\
    \bottomrule
	\end{tabular}
}{
 \caption{Text network datasets. C indicates the category, D is the total number of texts, Tr is the training set, Te is the test set, and N is the number of vertices of the graph network.}
 \label{tab1}
}
\capbtabbox{
 \begin{tabular}{lccc}
          \toprule
    Datasets & D & LR & Epoches    \\
    \midrule
    R52     & 0.6 & 0.005 & 950      \\
    OH     & 0.2 & 0.01 & 230       \\
    20NG     & 0.0 & 0.01 & 210   \\
    R8     & 0.2 & 0.005 & 300   \\
    MR     & 0.1 & 0.01 & 80   \\
    \bottomrule
	  \end{tabular}
}{
 \caption{The hyperparameters in our HLHG-3 model. D and LR denote dropout rate and learning rate respectively.}
 \label{tab2}
}
\end{floatrow}
\end{table*}

\paragraph{Baselines and experimental setting}We compare our HLHG with the following approaches: Convolutional Neural Network with pre-trained vectors (CNN-non-static)~\cite{Kim2014ConvolutionalNN}, LSTM model with pre-trained (LSTM-pretrain)~\cite{liu2016recurrent}, predictive text embedding for text classification (PTE)~\cite{tang2015pte}, fast text classifier (fastText)~\cite{Joulin2017BagOT}, simple word embedding models with simple pooling strategies (SWEM)~\cite{Shen2018BaselineNM}, label-embedding attentive models for text classification (LEAM)~\cite{Wang2018JointEO}, graph CNN model with Chebyshev filter (Graph-CNN-C)~\cite{defferrard2016convolutional}, graph CNN model with Spline filter (Graph-CNN-S)~\cite{bruna2013spectral}, graph CNN model with Fourier filter (Graph-CNN-F)~\cite{Henaff2015DeepCN}, and graph convolutional networks for text classification (text GCN)~\cite{yao2018graph}. The baseline models test by Yao et al~\cite{yao2018graph}.

In our HLHG-2 model, we set the dropout rate = 0.2. The learning rate is updated by Adam~\cite{kingma2014adam} during the training process. In our model, we set L2 loss weight as 0, we adopt early stopping. We set the learning rate to 0.02 for R8 dataset, and the remaining datasets uniformly set to 0.01. We set different epochs for different datasets. The epochs of R52 dataset is 350. The epochs of OH and 20NG dataset are 200, R8 and MR datasets are 60. In the HLHG-2 model, we set the number of hidden neurons of 1st convolutional layer as 128 for all datasets.

For our HLHG-3, we set the number of hidden neurons of the first convolutional layer to 128 except the MR dataset set to 64. In order to obtain better training results, we separately set different hyperparameters such as s dropout rate, learning rate, and epochs for different datasets. (See Table~\ref{tab2}) And the other parameters of HLHG-3 is the same as HLHG-2.

We construct the graph network for our HLHG-2 and HLHG-3 model, and the feature matrix and other parameters are same as Yao et al.~\cite{yao2018graph}.

\paragraph{Results}We show five datasets of supervised text classification accuracy in Table~\ref{tab3}. We demonstrate how our model performs on common splits taken from Yao et al.~\cite{yao2018graph}.

Table~\ref{tab3} presents classification accuracy and standard deviations of our models and the benchmark on the text network data. In general, our HLHG-2 and HLHG-3 achieve high levels of performance. Specifically, they achieve the best performance on R52,OH,20NG and R8. Compared to the best performance approach, the proposed models yield worse accuracies on the dataset MR. In general, the HLHG-3 and HLHG-2 models perform equally well. More specifically, the 3-order HLHG shows slightly better classification accuracy than the 2-order HLHG on most datasets. However, the difference in performance is not very large. Overall, the proposed architecture with hybrid high and low order neighborhood has good classification performance, which indicates that it not only effectively preserves the topological information of the graph, but also obtains a high-quality representation of the node.

\begin{table}
  \caption{Test Accuracy on text network classification. The values below the line are our methods. $ \pm $ represents the standard deviation of 100 runs with different random initializations.}
  \label{tab3}
  \centering
  \begin{tabular}{lccccc}
    \toprule
    Approaches & R52 & OH & 20NG & R8 & MR    \\
    \midrule
    CNN-non-static~\cite{Kim2014ConvolutionalNN} & 87.59 & 58.44 & 82.15 & 95.71 & \textbf{77.75}     \\
    LSTM-pretrain~\cite{liu2016recurrent} & 90.48 & 51.10 & 75.43 & 96.09 & 77.33  \\
    PTE~\cite{tang2015pte} & 90.71 & 53.58 & 76.74 & 96.69 & 70.23  \\
    fastText~\cite{Joulin2017BagOT} & 92.81 & 57.70 & 79.38 & 96.13 & 75.14     \\
    SWEM~\cite{Shen2018BaselineNM} & 92.94 & 63.12 & 85.16 & 95.32 & 76.65  \\
    LEAM~\cite{Wang2018JointEO} & 91.84 & 58.58 & 81.91 & 93.31 & 76.95  \\
    Graph-CNN-C~\cite{defferrard2016convolutional} & 92.75 & 63.86 & 81.42 & 96.99 & 77.22  \\
    Graph-CNN-S~\cite{bruna2013spectral} & 92.74 & 62.82 & - & 96.80 & 76.99     \\
    Graph-CNN-F~\cite{Henaff2015DeepCN} & 93.20 & 63.04 & - & 96.89 & 76.74  \\
    Text GCN~\cite{yao2018graph} & 93.56 & 68.36 & 86.34 & 97.07 & 76.74      \\
    \hline
    HLHG-2 (ours) & 94.21$\pm0.14$ & 69.16$\pm 0.19$ & \textbf{86.57$\pm$0.08} & \textbf{97.25$\pm$0.10} & 75.95$\pm0.14 $  \\
    HLHG-3 (ours) & \textbf{94.33$\pm$0.16} & \textbf{69.36$\pm$0.24} & 86.35$\pm$0.24 & \textbf{97.25$\pm$0.12} & 76.49$\pm0.32$  \\
    \bottomrule
  \end{tabular}
\end{table}

Table~\ref{tab4} shows the Comparison of network complexity and quantity of parameter with the Text GCN~\cite{yao2018graph}. Our HLHG can match with the Text GCN on computational complexity, and less than Text GCN on quantity of parameter. As described in subsection 3.3, the number of features in the dataset node is much larger than the number of neurons in the hidden convolutional layer. Therefore, we only compare the computational complexity and number of parameters of the first convolutional layer in our HLHG model. In the Table~\ref{tab4}, Comp.and Params represent the computational complexity and parameter quantities of the first-layer of the graph convolutional network, respectively. In the results of the computational complexity, the first constant denotes the neurons number of the first convolutional layer, and the second constant denotes the order of adjacency matrix. The parameter $ m $ denotes the number of non-zero entries of the sparse regularization adjacency matrix. And the parameter $ r $ denotes the feature dimension of node in graph network.

In Text GCN~\cite{yao2018graph}, the number of hidden neurons of the first convolutional layer is 200, therefore complexity and params has the constant 200. In our HLHG-2 model, the constant 128 denotes the number of hidden neurons of the first convolutional layer. And the constant 2 means the highest order of HLHG-2. In our HLHG-3 model, the constant 128 and 64 denote the number of hidden neurons of the first convolutional layer. The constant 3 represents the highest order of the corresponding model. The result in the Table~\ref{tab4} show, our HLHG-3 model has an advantage in computational complexity in dataset MR. Because of the weight share in different order neighborhood, our HLHG models require less trainable weight parameters. Especially on dataset MR, the parameter amount is only 1/3 of Text GCN~\cite{yao2018graph}.

\begin{table}
  \caption{Comparison of network complexity and quantity of parameters. Comp. and Params represent the computational complexity and parameter quantities of the first-layer of the graph convolutional network, respectively. The first constant of Comp. and Params. present the the number of hidden neurons of the first convolutional layer. The second constant of Comp. denotes the order of the neighborhood matrix.}
  \label{tab4}
  \centering
  \begin{tabular}{lcc}
    \toprule
    Approaches & Comp. & Params    \\
    \midrule
    Text GCN~\cite{yao2018graph} & \textbf{O(200$\times$ 1 $\times$ m $\times$ r)} & $ O (200\times r) $     \\
    \hline
    HLHG-2 (ours) & $ O (128\times 2\times m \times r) $ & \textbf{O(128 $\times$ r)} \\
    HLHG-3 (ours) & \textbf{O(64 $\times$ 3 $\times$ m $\times$ r)} (MR dataset)  & \textbf{O(64 $\times$ r)} (MR dataset)   \\
    HLHG-3 (ours) & $ O (128\times 3\times m \times r) $ (other datasets) & \textbf{O(128 $\times$ r)} (other datasets)  \\
    \bottomrule
  \end{tabular}
\end{table}

\subsection{semi-supervised node classification}

\paragraph{Datasets}In the semi-supervised node classification, we use the citation network datasets, Citeseer, Cora and Pubmed~\cite{sen2008collective}. In these citation datasets, the nodes represent the article published in the corresponding journal. The edges between the two nodes represent references from one article to another, and the tags represent the topic of the article. The citation link constructs an adjacency matrix. The summary statistics features of citation graph are shown in Table~\ref{tab5}.

\begin{table*}
\begin{floatrow}
\capbtabbox{
 \begin{tabular}{lcccccc}
	\toprule
    Datasets & N & E & F & LR & C    \\
    \midrule
    Cora     & 2708 & 5429 & 1433 & 0.052 & 7     \\
    Citeseer     & 3327 & 4732 & 3703 & 0.036 & 6      \\
    Pubmed     & 19717 & 44338 & 500 & 0.003 & 3  \\
    \bottomrule
	\end{tabular}
}{
 \caption{Citation network datasets. N means the node of the citation, E means the edge of the citation, and F means the feature of the nodes. LR and C indicate the learning rate and category, respectively.}
 \label{tab5}
}
\capbtabbox{
 \begin{tabular}{lcccc}
          \toprule
    Datasets & D & LR & ES & Ep    \\
    \midrule
    Cora     & 0.5 & 0.01 & no & 500      \\
    Citeseer  & 0.5 & 0.005 & 5 & 500       \\
    Pubmed    & 0.6 & 0.01 & 1 & 200   \\
    \bottomrule
	  \end{tabular}
}{
 \caption{The hyperparameters of HLHG-3. D, LR, ES and Ep denote dropout rate, learning rate, early stopping and epoches, respectively.}
 \label{tab6}
}
\end{floatrow}
\end{table*}

\paragraph{Baselines and experimental setting}We compare our HLHG with the same baseline methods as in Abu-El-Haija et al.~\cite{abu2019mixhop} and Yang et al.~\cite{Yang2016RevisitingSL}. The baselines are determined as follows: manifold regularization (ManiReg)~\cite{Belkin2006ManifoldRA}, semi-supervised embedding (SemiEmb)~\cite{Weston2008DeepLV}, label propagation (LP)~\cite{Zhu2003SemiSupervisedLU}, skip-gram based graph embeddings (DeepWalk)~\cite{Perozzi2014DeepWalkOL}, iterative classification algorithm (ICA)~\cite{Lu2003LinkbasedC}, Planetoid~\cite{Yang2016RevisitingSL}, HO~\cite{abu2018a}, MixHop~\cite{abu2019mixhop}.

For HLHG-2, we use the following sets on citation datasets (Cora, Citeseer, and Pubmed): 16 (number of hidden units), 0.5 (dropout rate), 0.0005 (L2 regularization),10(early stopping), 300(epochs) and 0.01(learning rate).

For HLHG-3 model, we set different numbers of hidden neurons for different datasets. We set 8 hidden neurons for Citeseer dataset to reduce computational complexity and parameter quantities, whereas set 10 hidden neurons for Cora and Pubmed datasets to capture richer features. The hyperparameters of HLHG-3 set as Table~\ref{tab6}.

\paragraph{Results}In the semi-supervised experiments, we train and test our models on those citation network datasets follow the methodology proposed in Yang et al.~\cite{Yang2016RevisitingSL}. The classification accuracy is the mean of 100 runs with random weight initializations.

Table~\ref{tab7}, the node classification accuracy values above line are copied from Abu-El-Haija~\cite{abu2018a,abu2019mixhop} and Yang et al.~\cite{Yang2016RevisitingSL}. The values below the line are our HLHG models. $ \pm $ represents the standard deviation of 100 runs with different random initializations. These splits utilize only 20 labeled nodes per class during training. We achieve the best test accuracy of 82.7\%, 71.5\%, and 79.3\% on Cora, Citeseer, and Pubmed respectively. Compared with other high order graph convolutional neural network~\cite{abu2018a,abu2019mixhop} on the same datasets, they get the high order information by linear combinations of features from farther distances. Our HLHG model is nonlinear to get the high order neighborhood information.

\begin{table}
  \caption{Test Accuracy on citation network classification. The benchmark test result copy from~\cite{abu2019mixhop} and~\cite{Yang2016RevisitingSL}. The mean standard deviation of our model is the average of 100 times runs.}
  \label{tab7}
  \centering
  \begin{tabular}{lccccc}
    \toprule
    Methods & Cora & Citeseer & Pubmed    \\
    \midrule
    ManiReg~\cite{Belkin2006ManifoldRA} & 59.5 & 60.1 & 70.7     \\
    SemiEmb~\cite{Weston2008DeepLV} & 59.0 & 59.6 & 71.1     \\
    LP~\cite{Zhu2003SemiSupervisedLU} & 68.0 & 45.3 & 63.0     \\
    DeepWalk~\cite{Perozzi2014DeepWalkOL} & 67.2 & 43.2 & 65.3     \\
    ICA~\cite{Lu2003LinkbasedC} & 75.1 & 69.1 & 73.9     \\
    Planetoid~\cite{Yang2016RevisitingSL} & 75.7 & 64.7 & 77.2     \\
    GCN~\cite{kipf2016semi} & 81.5 & 70.3 & 79.0     \\
    \hline
    HO-3~\cite{abu2018a} & 81.6$ \pm 0.47 $ & 71.2$\pm 0.94 $ & 80.0$ \pm 0.64 $   \\
    HO-4~\cite{abu2018a} & 81.6$ \pm 0.63 $ & 71.2$\pm 0.84 $ & 80.1$ \pm 0.65 $   \\
    MixHop~\cite{abu2019mixhop} & 81.8$ \pm 0.62 $ & 71.4$\pm 0.81 $ & 80.0$ \pm 1.10$   \\
    MixHop (learned)~\cite{abu2019mixhop} & 81.9$ \pm 0.40 $ & 71.4$\pm 0.81 $ & \textbf{80.8$\pm$0.58}  \\
    \hline
    HLHG-2 (ours) & \textbf{82.7$\pm$0.28} & \textbf{71.5$\pm$0.22} & 79.1$ \pm 0.18 $   \\
    HLHG-3 (ours) & \textbf{82.7$\pm$0.29} & \textbf{71.5$\pm$0.39} & 79.3$ \pm 0.15 $   \\
    \bottomrule
  \end{tabular}
\end{table}

In Table~\ref{tab8}, we compare the network complexity and quantity of parameter with the high order graph convolutional network and the classic GCN. The result shows our model matches with the other approaches in the computational complexity. In the parameter quantities, our HLHG-3 model is less than the GCN~\cite{kipf2016semi}. The reason is that our model shares the weigh in the same layer among the different order neighborhood matrix.

\begin{table}
  \caption{Comparison of network complexity and quantity of parameters. Comp. and Params represent the computational complexity and parameter quantities of the first-layer of the graph convolutional network, respectively. The second constant number represents the order number neighborhood matrix.}
  \label{tab8}
  \centering
  \begin{tabular}{lcc}
    \toprule
    Approaches & Comp. & Params    \\
    \midrule
    GCN~\cite{kipf2016semi} & \textbf{O(16 $\times$ 1 $\times$ m $\times$ r)} & $ O (16\times 1\times r) $     \\
    \hline
    HO-3~\cite{abu2018a} & $ O (10\times 3\times m\times r) $ & $ O (10\times 3\times r) $     \\
    HO-4~\cite{abu2018a} & $ O (10\times 4\times m\times r) $ & $ O (10\times 4\times r) $     \\
    MixHop~\cite{abu2019mixhop} & $ O (20\times 2\times m\times r) $ & $ O (20\times 3\times r) $     \\
    MixHop (learned)~\cite{abu2019mixhop} & $ O (20\times 2\times m\times r) $ & $ O (60\times 1\times r) $     \\
    \hline
    HLHG-2 (ours) & $ O (16\times 2\times m \times r) $ & $ O (16\times 1\times r) $ \\
    HLHG-3 (ours) & $ O (8\times 3\times m \times r) $ (Citeseer) & \textbf{O(8 $\times$ 1 $\times$ r)} (Citeseer) \\
    HLHG-3 (ours) & $ O (10\times 3\times m \times r) $ (other) & \textbf{O(10 $\times$ 1 $\times$ r)} (other) \\
    \bottomrule
  \end{tabular}
\end{table}

\section{Conclusion}

In this paper, we propose a hybrid low order and higher order GCN model for supervised classification on the text network dataset and for semi-supervised classification on the citation network. In our model, we propose a novel information fusion layer which is nonlinear to combine the low and higher order neighborhood. To reduce the parameter, we propose to share the weigh in the same convolutional layer for different order neighborhood. Experiments on the two sets network datasets suggest the HLHG mode has the capability to fuse higher order neighborhood for supervised classification and semi-supervised classification. Our model outperforms significant performance than the benchmark. We also analyze the computational complexity and parameter quantity less than high order method. For future work, we will extend our model to fuse with graph attention networks~\cite{Velickovic2018GraphAN}.

{\small
\bibliographystyle{ieee_fullname}
\bibliography{egbib}
}

\end{document}